\documentclass[10pt,twocolumn,letterpaper]{article}

\usepackage{cvpr}
\usepackage{booktabs}
\usepackage{graphicx}
\usepackage[table]{xcolor}

\definecolor{cvprblue}{rgb}{0.21,0.49,0.74}
\usepackage[pagebackref,breaklinks,colorlinks,allcolors=cvprblue]{hyperref}

\def\paperID{*****}
\def\confName{CVPR}
\def\confYear{2026}

\title{RANGER: Sparsely-Gated Mixture-of-Experts with Adaptive Retrieval Re-ranking for Pathology Report Generation}

\author{
Yixin Chen$^{1}$ \quad
Ziyu Su$^{2}$ \quad
Hikmat Khan$^{2}$ \quad
M.~Khalid Khan Niazi$^{1,2}$ \\[0.5em]
$^{1}$Department of Biomedical Engineering, The Ohio State University, Columbus, OH, USA \\
$^{2}$Department of Pathology, The Ohio State University, Columbus, OH, USA
}

\begin{document}
\maketitle

\begin{abstract}
Pathology report generation remains a relatively under-explored downstream task, primarily due to the gigapixel scale and complex morphological heterogeneity of Whole Slide Images (WSIs). Existing pathology report generation frameworks typically employ transformer architectures, relying on a homogeneous decoder architecture and static knowledge retrieval integration. Such architectures limit generative specialization and may introduce noisy external guidance during the report generation process. To address these limitations, we propose RANGER, a sparsely-gated Mixture-of-Experts (MoE) framework with adaptive retrieval re-ranking for pathology report generation. Specifically, we integrate a sparsely gated MoE into the decoder, along with noisy top-$k$ routing and load-balancing regularization, to enable dynamic expert specialization across various diagnostic patterns. Additionally, we introduce an adaptive retrieval re-ranking module that selectively refines retrieved memory from a knowledge base before integration, reducing noise and improving semantic alignment based on visual feature representations. We perform extensive experiments on the PathText-BRCA dataset and demonstrate consistent improvements over existing approaches across standard natural language generation metrics. Our full RANGER model achieves optimal performance on PathText dataset, reaching BLEU-1 to BLEU-4 scores of 0.4598, 0.3044, 0.2036, and 0.1435, respectively, with METEOR of 0.1883, and ROUGE-L of 0.3038, validating the effectiveness of dynamic expert routing and adaptive knowledge refinement for semantically grounded pathology report generation.

\end{abstract}

\section{Introduction}
\label{sec:intro}

The automated analysis of Whole Slide Images (WSIs) has become an essential focus in computational pathology (CPath). The digital pathology workflows and advances in deep learning-based frameworks \cite{lu2021data, shao2021transmil, chen2026histomet, chen2024towards} significantly advances in histopathological artificial intelligence. Among a variety of downstream tasks, generating pathology report \cite{guo2024histgen, zhang2025historical, lucassen2025pathology, hu2025pathology} from whole slide images directly is one of the most clinically meaningful tasks, as it can reduce pathologists workload and assist clinical decision making. However, the ultra-high resolution and complex morphological heterogeneity of WSIs pose substantial challenges for learning semantically meaningful and diagnostically relevant representations.\par

Although there are significant amount of advancement in image captioning and medical report generation \cite{jin2026grounded, gao2025s2d, che2025llm, wang2025activating, ma2025fully}
, generating diagnostic reports directly from whole slide images remains substantially more challenging. WSIs are gigapixel-scale images containing thousands of heterogeneous tissue patches, of which only a small fraction are diagnostically informative. Effectively extracting semantically meaningful and clinically relevant features from such ultra-high resolution data is non-trivial. Existing report generation framework typically relying on transformer \cite{vaswani2017attention} structure, where a homogeneous decoder is employed to generate diagnostic narrative. While such models can capture global contextual dependencies, they rely on a single shared set of parameters to model diverse pathological patterns and linguistic expressions. However, pathology reports often involve heterogeneous reasoning processes, ranging from morphological description to grading, staging, and biomarker interpretation, which may benefit from specialized generative behaviors. Moreover, although knowledge-enhanced methods have been explored to incorporate external knowledge to augment the report generation process, retrieved information is often directly fused without adaptive selection or refinement. Therefore, the retrieval process might introduce noise or weak relevant content, which will affect the downstream decoder performance in generating diagnostic reports. Although these models can capture contextual dependencies, their design inevitably (1) limits generative specialization due to a homogeneous decoder architecture and (2) introduces noisy or weakly relevant information through static knowledge integration. \par

To address these limitations, We introduce a memory-augmented, sparsely-gated Mixture-of-Experts (MoE) framework with noisy routing and load-balancing regularization to enable dynamic and stable expert specialization for pathology report generation. for pathology report generation that introduces decoder-level specialization and adaptive knowledge refinement. Our MOE decoder enables heterogeneous diagnostic reasoning by allowing multiple experts to model distinct morphological and linguistic patterns. Meanwhile, an adaptive retrieval re-ranking module selectively refines retrieved knowledge before integration, mitigating noisy guidance and improving semantic grounding. Extensive experiments on the PathText-BRCA dataset, one of the latest pathology report generation dataset from TCGA, demonstrate consistent improvements over existing methods in terms of natural language processing (NLP) metrics. 

\begin{itemize}
    \item \textbf{Adaptive Retrieval Re-ranking.}
    We propose a two-stage retrieval framework that refines retrieved textual knowledge before decoder integration, suppressing noisy guidance and improving semantic grounding for pathology report generation.

    \item \textbf{Sparsely-Gated Mixture-of-Experts Decoder.}
    We introduce a sparsely-gated MoE decoder that enables dynamic expert specialization for heterogeneous morphological and linguistic patterns, enhancing conditional generative modeling without proportional computational overhead.

    \item \textbf{Performance on Public Dataset. }
    On the PathText-BRCA dataset, our full model achieves a BLEU-4 score of \textbf{0.1435} and ROUGE-L of \textbf{0.3038}, demonstrating consistent improvements over strong report-generation baselines and validating the effectiveness of adaptive knowledge refinement and expert specialization.
\end{itemize}

\begin{figure*}[t]
    \centering
    \includegraphics[width=\textwidth]{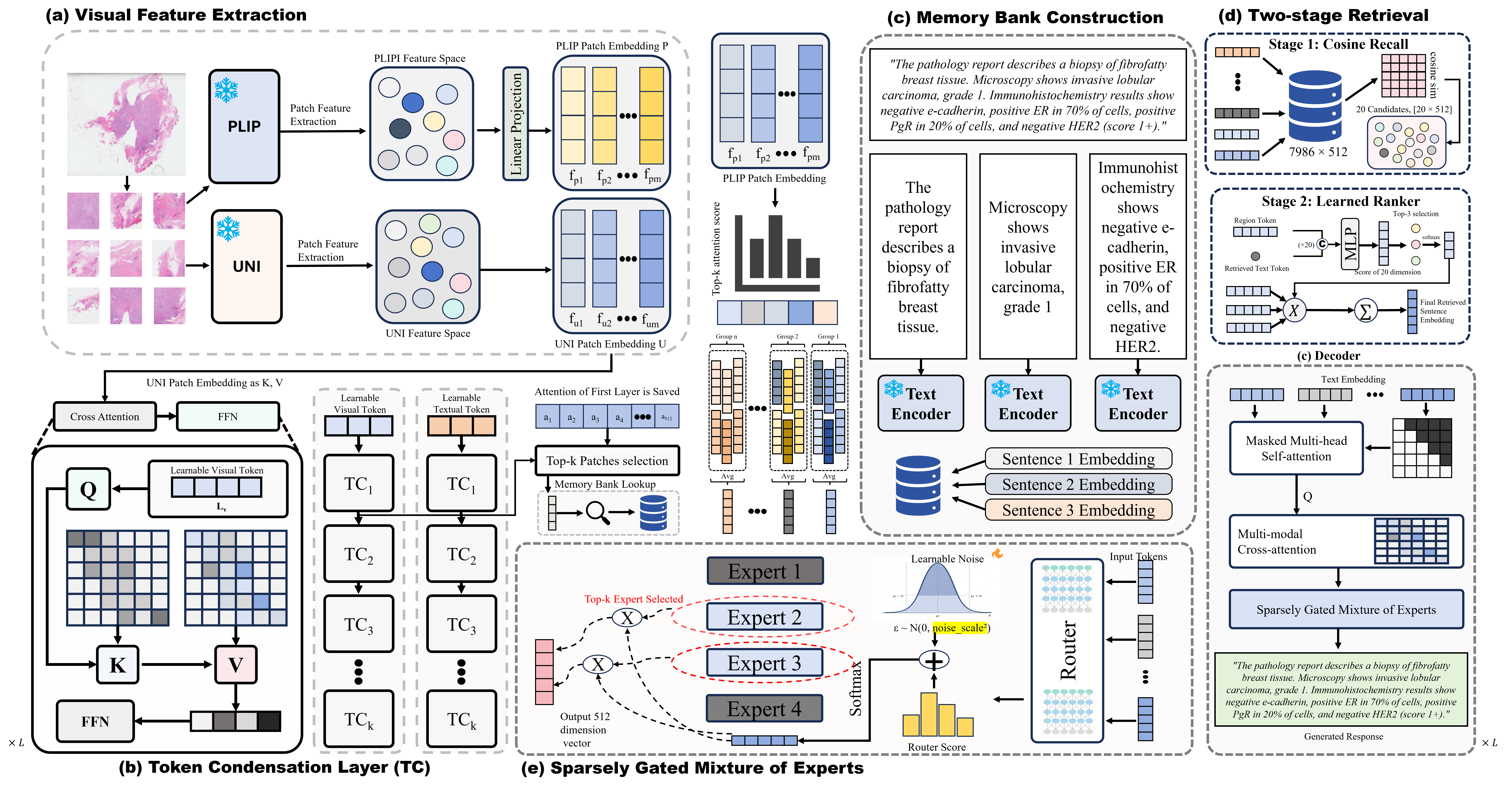}
    \caption{
\textbf{Overview of the proposed RANGER framework.}
(a) Visual branch for whole-slide image feature extraction. 
(b) Dual learnable tokens that cross-attend to visual features and retrieved textual knowledge, respectively, enabling cross-modal interaction. 
(c) Construction of a sentence-level memory bank from historical pathology reports. 
(d) Two-stage retrieval module that first performs similarity-based candidate selection and then applies a learned re-ranking mechanism for adaptive knowledge refinement. 
(e) Sparsely-gated Mixture-of-Experts (MoE) decoder with noisy top-$k$ routing and load-balancing regularization for conditional expert specialization during report generation.
}
    \label{fig:framework}
\end{figure*}


\section{Related Work}
\label{related_work}

\subsection{Pathology Report Generation from Whole-Slide Images}

Automated pathology report generation extends image captioning techniques to ultra-high-resolution whole-slide images (WSIs). Unlike natural images, WSIs contain thousands of tissue patches, resulting in extreme spatial redundancy and substantial morphological heterogeneity. Early approaches adapt attention-based CNN--RNN \cite{vinyals2015show} or Transformer \cite{vaswani2017attention} architectures to aggregate patch-level features for textual generation. Sengupta \textit{et al.}~\cite{sengupta2024automatic} achieve end-to-end pathology report generation directly from whole-slide images. Quilt-LLaVA~\cite{seyfioglu2024quilt} enables spatially grounded WSI reasoning for visual question answering and pathology report generation. Methods such as MI-Gen \cite{chen2024wsicaption} and HistGen \cite{guo2024histgen} introduce multiple-instance or hierarchical encoding strategies to capture local--global interactions across WSI regions. 

Despite these advances, purely visual modeling often lacks explicit semantic grounding and remains susceptible to redundancy introduced by non-diagnostic tissue regions. Furthermore, aggregating massive patch embeddings into a compact representation remains challenging due to the diverse morphological patterns present across pathological cases.

\subsection{Vision\text{-}Language Modeling and Knowledge-Enhanced Pathology Report Generation}

Recent advances in pathology report generation have increasingly leveraged vision–language modeling to enable end-to-end learning from whole-slide images (WSIs). Polypath~\cite{ahmed2025polypath} adapts a large multimodal model to generate cohesive diagnostic reports from multiple WSIs within a unified framework. Lucassen \textit{et al.}~\cite{lucassen2025pathology} explore multimodal representation learning for pathology report generation in cutaneous melanocytic lesions. Hu \textit{et al.}~\cite{hu2025pathology} incorporate knowledge retrieval with multi-level regional feature selection to support report generation from WSIs. In parallel, Chen \textit{et al.}~\cite{chen2025evidence} introduce a multi-agent diagnostic system that collaborates to generate comprehensive pathology reports through evidence-based reasoning.

To better reflect clinical reasoning processes, recent approaches further integrate historical textual knowledge into medical report generation. Retrieval-augmented frameworks leverage memory modules or sentence-level knowledge banks to enhance factual consistency and semantic completeness. BiGen~\cite{zhang2025historical} proposes a knowledge-guided bi-modal concurrent learning framework that constructs a sentence-level knowledge bank and retrieves WSI-relevant textual features via similarity-based matching. The retrieved knowledge is integrated through cross-attention mechanisms with shared weights to promote cross-modal alignment between visual and textual representations, enriching semantic representations while suppressing redundant information. However, the transformation applied to aggregated global tokens remains homogeneous across samples, implicitly assuming identical transformation dynamics across diverse pathological regimes.

\subsection{Mixture-of-Experts for Conditional Modeling}

Mixture-of-Experts (MoE) introduces input-dependent conditional computation to substantially increase model capacity without proportionally increasing computational cost. In transformer architectures, MoE replaces the standard feed-forward network with a collection of expert-specific feed-forward networks, and it typically employs a gating function that dynamically routes each input to a subset of experts. Large-scale implementations such as GLaM~\cite{du2022glam}, ST-MoE~\cite{zoph2022st}, and OpenMoE~\cite{xue2024openmoe} demonstrate that sparse expert activation enables scalable parameter growth while maintaining training and inference efficiency.

Sparse MoE typically employs top-$k$ routing to activate a small number of experts per token, often augmented with auxiliary load-balancing objectives to mitigate expert under-utilization~\cite{lewis2021base}. To address limitations of discrete routing, recent works propose differentiable gating mechanisms such as DSelect-$k$~\cite{hazimeh2021dselect}, token-adaptive routing strategies as in AdaMoE~\cite{zeng2024adamoe}, and fully differentiable expert merging formulations including Lory~\cite{zhong2024lory} and soft merging approaches~\cite{muqeeth2023soft}. These advances improve routing stability, expert specialization, and scalability across diverse modeling regimes.

By enabling piecewise transformations conditioned on input representations, MoE reduces interference across heterogeneous data distributions and promotes regime-specific specialization. While MoE has achieved strong empirical success in large language models and multimodal transformers, its application to pathology report generation remains largely unexplored. Given the substantial morphological and semantic heterogeneity inherent in whole-slide images (WSIs), conditional transformation of aggregated visual and textual representations provides a principled mechanism for enhancing regime-specific modeling while preserving cross-modal alignment within knowledge-guided report generation frameworks.

\section{Method}
\label{sec:method}

\subsection{Problem Formulation}
\label{subsec:problem}

We formulate pathology report generation from whole slide images (WSIs) as conditional sequence modeling over a set-structured visual representation. 
A WSI is encoded as a collection of visual tokens 
\(
\mathcal{V} = \{ \mathbf{v}_i \}_{i=1}^{|\mathcal{V}|}, 
\mathbf{v}_i \in \mathbb{R}^{d},
\)
where each token represents a spatially localized tissue region extracted by a pretrained visual encoder. 
Due to the gigapixel scale and morphological variability of WSIs, the visual input is inherently unordered and highly redundant. 
The target diagnostic report is modeled as an ordered token sequence 
\(
\mathcal{Y} = (y_1, \dots, y_T),
\)
which captures hierarchical pathological reasoning ranging from low-level morphology to high-level diagnostic interpretation. 
The generation process is modeled autoregressively:

\begin{equation}
P(\mathcal{Y} \mid \mathcal{V}) 
= \prod_{t=1}^{T} 
P(y_t \mid \mathcal{Y}_{<t}, \mathcal{V}),
\end{equation}

This formulation presents intrinsic structural challenges. 
First, diagnostically informative evidence is sparsely distributed among a large number of visual tokens, requiring selective grounding of relevant morphology. 
Second, different segments of the report exhibit heterogeneous generative patterns, implying that distinct semantic transformations may be needed during decoding. 
The model is trained by minimizing the negative log-likelihood:

\begin{equation}
\mathcal{L}_{\text{gen}} 
= - \sum_{t=1}^{T} 
\log P(y_t \mid \mathcal{Y}_{<t}, \mathcal{V}).
\end{equation}


\subsection{Framework Overview}
\label{subsec:overview}

Figure~\ref{fig:framework} illustrates the overall architecture.
Given a WSI, patch-level visual features are first extracted by pretrained encoders 
and projected into a shared embedding space. 
These visual tokens are spatially grouped to form region-level representations, 
which serve as a semantic anchor for memory grounding.

For each region token, we perform a two-stage knowledge retrieval process. 
A cosine-based coarse recall retrieves candidate sentence embeddings 
from a sentence-level memory bank. 
A learned reranker then refines these candidates by modeling 
fine-grained region–text compatibility, producing region-aware textual embeddings. 
The refined textual features are fused with visual representations 
before being fed into the decoder.

To model heterogeneous generative patterns in pathology reports, 
we replace the standard Transformer feed-forward network with a 
sparsely-gated Mixture-of-Experts (MoE) module. 
An input-dependent router selects a subset of experts for each token, 
enabling conditional computation and token-level specialization. 

Overall, the framework jointly performs 
(i) adaptive semantic grounding via retrieval refinement, and 
(ii) transformation specialization via sparse expert routing,
thereby addressing both input sparsity and output heterogeneity.

\subsection{Token Condensation Layer}
\label{subsec:tc}

To condense dense patch-level features into compact semantic summaries,
we introduce a Token Condensation (TC) layer in both visual and textual branches.

\paragraph{Visual Branch.}

Given UNI patch embeddings 
\(
\{ f^{u}_1, \dots, f^{u}_m \},
\)
we introduce a single learnable visual token 
\(
\ell_v \in \mathbb{R}^{d}.
\)

The condensation process is implemented as cross-attention,
where the learnable token serves as query and patch embeddings serve as keys and values:

\begin{equation}
h_v = \text{CrossAttn}(Q=\ell_v, K=\{f^{u}_i\}, V=\{f^{u}_i\}).
\end{equation}

The output is further transformed by a feed-forward network with residual connection:

\begin{equation}
z_v = \text{FFN}(h_v) + h_v.
\end{equation}

The resulting token $z_v$ summarizes global morphological information
from all patches.

\paragraph{Text Branch.}

Similarly, given PLIP sentence embeddings 
\(
\{ f^{p}_1, \dots, f^{p}_M \},
\)
we introduce a learnable textual token 
\(
\ell_t \in \mathbb{R}^{d}.
\)

Cross-attention is applied to the retrieved top-k sentence embeddings:

\begin{equation}
h_t = \text{CrossAttn}(Q=\ell_t, K=\{f^{p}_i\}, V=\{f^{p}_i\}),
\end{equation}

followed by:

\begin{equation}
z_t = \text{FFN}(h_t) + h_t.
\end{equation}

This single-token bottleneck enables global aggregation while maintaining
differentiability and cross-modal compatibility.


\subsection{Adaptive Knowledge Retrieval}
\label{subsec:retrieval}

\paragraph{Memory Bank Construction.}
We construct a fixed memory bank 
$\mathcal{B} = \{ \mathbf{b}_i \}_{i=1}^{M}$, 
where each $\mathbf{b}_i \in \mathbb{R}^{d}$ 
is a PLIP-encoded embedding of a sentence extracted from training pathology reports. All sentence embeddings are pre-computed 
using the PLIP text encoder and remain frozen during training. The memory bank serves as an auxiliary semantic repository, providing textual evidence for visual region representations.

\paragraph{Memory Retrieval.}
Given a whole-slide image, we first encode all patches 
$\{ w_i \}_{i=1}^{N}$ using the PLIP image encoder, obtaining patch embeddings $\{ \mathbf{f}_{pi} \}_{i=1}^{N}$, where $\mathbf{f}_{pi} \in \mathbb{R}^{d}$. Obviously, retrieving auxiliary semantic memory for all patches are computationally heavy, as many regions may be diagnostically irrelevant. Following the attention-guided token selection strategy adopted in the Bigen framework~\cite{zhang2025historical},
we utilize attention scores from the first visual cross-attention layer to identify salient patches.

Specifically, we retain the top-$K$ patches 
with the highest attention weights, 
forming a reduced set 
$\mathcal{P} = \{ \mathbf{f}_{pi} \}_{i \in \mathcal{I}_{\text{top}}}$, 
where $|\mathcal{I}_{\text{top}}| = K$.

To further reduce redundancy while preserving spatial coherence of tissue structures, we uniformly partition $\mathcal{P}$ 
into groups of size $m$, 
yielding $R = K/m$ region tokens. 
Each region embedding is obtained via average pooling:

\begin{equation}
\bar{\mathbf{f}}_r =
\frac{1}{m}
\sum_{i \in \mathcal{G}_r}
\mathbf{f}_{pi},
\quad r = 1, \dots, R.
\end{equation}

The resulting region embeddings 
$\{ \bar{\mathbf{f}}_r \}_{r=1}^{R}$ 
serve as queries for subsequent memory retrieval.

\paragraph{Stage 2: Learned Re-ranking.}
To model higher-order cross-modal alignment, 
we introduce a parametric compatibility function. 
For each region embedding 
$\bar{\mathbf{f}}_r$ 
and each retrieved candidate 
$\mathbf{b}_i \in \mathcal{C}_r$, 
we compute:

\begin{equation}
s_{r,i} = 
\text{MLP}\left(
[\bar{\mathbf{f}}_r \,\|\, \mathbf{b}_i]
\right),
\end{equation}

where $\text{MLP}: \mathbb{R}^{2d} \rightarrow \mathbb{R}$ 
is a learnable scoring function, 
and $[\cdot \| \cdot]$ denotes vector concatenation.

To further focus on the most compatible evidence, 
we retain the top-$k$ candidates 
according to $s_{r,i}$ and compute a normalized aggregation:

\begin{equation}
\tilde{\mathbf{f}}_r =
\sum_{i \in \mathcal{I}_r}
\frac{\exp(s_{r,i})}
{\sum_{i' \in \mathcal{I}_r} \exp(s_{r,i'})}
\, \mathbf{b}_i,
\quad |\mathcal{I}_r| = k,
\end{equation}

where $\mathcal{I}_r$ denotes the indices of the 
top-$k$ re-ranked candidates (we set $k=3$). Thus, this learned re-ranking module captures non-linear interactions between visual morphology and textual semantics. 
By jointly optimizing the reranker 
with the generation objective, 
the retrieval process becomes task-aware and fully differentiable.


\subsection{Sparsely-Gated MoE Decoder}
\label{subsec:moe}

Standard Transformer decoders typically apply an identical FFN to every token. Suppose h denotes the token representation after attention:

\begin{equation}
\mathbf{h}' = \text{FFN}(\mathbf{h}),
\end{equation}

which assumes a shared transformation suffices for diverse 
linguistic structures. However, pathology reports exhibit 
heterogeneous token distributions and reasoning trajectories. Therefore, we replace each FFN with a Mixture-of-Experts module:

\begin{equation}
\text{MoE}(\mathbf{h}) =
\sum_{e=1}^{E} p_e(\mathbf{h}) f_e(\mathbf{h}),
\end{equation}

where $\{ f_e \}_{e=1}^{E}$ are expert FFNs ($E=4$).

\paragraph{Noisy Top-$k$ Routing.}
For each hidden representation token, the router computes logits by adding a noise using Softplus, which ensures positive and learnable scale for the added noise. The noise is injected during training to prevent the expert collapse scenario:

\begin{equation}
g(\mathbf{h}) =
W_r \mathbf{h}
+ \epsilon \odot \text{Softplus}(W_n \mathbf{h}),
\quad \epsilon \sim \mathcal{N}(0, I).
\end{equation}

We select top-$k=2$ experts and compute sparse probabilities for each expert by using softmax:

\begin{equation}
p_e(\mathbf{h}) =
\frac{\exp(g_e)}
{\sum_{e' \in \text{Top-}k} \exp(g_{e'})}
\mathbf{1}[e \in \text{Top-}k].
\end{equation}

The noise will ensure the expert collapse, there for each expert will have gradients to update itself during training.

\paragraph{Specialization Perspective.}
This sparse routing mechanism allows different experts 
to implicitly specialize in distinct linguistic patterns, 
such as morphological descriptors, quantitative assessments, 
or bio-marker related findings, 
thereby increasing expressive capacity without linear scaling 
of computational cost.

\paragraph{Load Balancing.}
To prevent expert under-utilization, 
we adopt the auxiliary loss:

\begin{equation}
\mathcal{L}_{\text{aux}} =
E \sum_{e=1}^{E}
f_e^{\text{(usage)}} \cdot p_e^{\text{(mean)}}.
\end{equation} where $f_e^{\text{(usage)}}$ is the fraction of tokens dispatched to expert $e$ (hard routing indicator), and $p_e^{\text{(mean)}}$ is the mean soft router probability assigned to expert $e$, both averaged over all tokens in the batch. This regularizer encourages uniform expert utilization 
and stabilizes training.


\subsection{Training Objective}
\label{subsec:training}

The final objective is:

\begin{equation}
\mathcal{L} =
\mathcal{L}_{\text{LM}}
+ \lambda \mathcal{L}_{\text{aux}},
\quad \lambda = 0.01.
\end{equation}

All parameters, including the reranker and router, 
are optimized end-to-end. 
This unified optimization ensures that knowledge refinement 
and expert specialization co-adapt to maximize 
diagnostic generation quality.

\section{Experiment}
\label{experiment}

\begin{table*}[t]
\centering
\caption{Results of pathology report generation on PathText (BRCA).
BLEU-n denotes n-gram BLEU score. $\uparrow$ indicates higher is better.
\textbf{Bold} denotes the best result.}
\label{tab:pathtext_brca_results}

\setlength{\tabcolsep}{6pt}
\renewcommand{\arraystretch}{1.2}

\begin{tabular*}{\textwidth}{@{\extracolsep{\fill}}lcccccc}
\toprule
Model 
& BLEU-1 $\uparrow$ 
& BLEU-2 $\uparrow$ 
& BLEU-3 $\uparrow$ 
& BLEU-4 $\uparrow$ 
& METEOR $\uparrow$ 
& ROUGE-L $\uparrow$ \\
\midrule
CNN-RNN \cite{vinyals2015show}             
& 0.371 & 0.185 & 0.089 & 0.043 & 0.143 & 0.239 \\

att-LSTM \cite{xu2015show}                
& 0.371 & 0.191 & 0.094 & 0.048 & 0.142 & 0.238 \\

Vanilla Transformer \cite{vaswani2017attention} 
& 0.389 & 0.246 & 0.157 & 0.103 & 0.158 & 0.257 \\

R2Gen \cite{chen2020generating}           
& 0.378 & 0.243 & 0.160 & 0.107 & 0.179 & 0.279 \\

R2GenCMN \cite{chen2021cross}             
& 0.396 & 0.254 & 0.164 & 0.110 & 0.163 & 0.279 \\

MI-Gen \cite{chen2024wsicaption}          
& 0.416 & 0.267 & 0.174 & 0.115 & 0.165 & 0.270 \\

HistGen \cite{guo2024histgen}             
& 0.422 & 0.272 & 0.177 & 0.118 & 0.169 & 0.277 \\

BiGen \cite{zhang2025historical}          
& 0.450 & 0.296 & 0.196 & 0.135 & 0.180 & 0.293 \\
\midrule
\textbf{RANGER (ours)} 
& \textbf{0.4598} 
& \textbf{0.3044} 
& \textbf{0.2036} 
& \textbf{0.1435} 
& \textbf{0.1883} 
& \textbf{0.3038} \\
\bottomrule
\end{tabular*}
\end{table*}

\begin{figure*}[t]
    \centering
    \includegraphics[width=\textwidth]{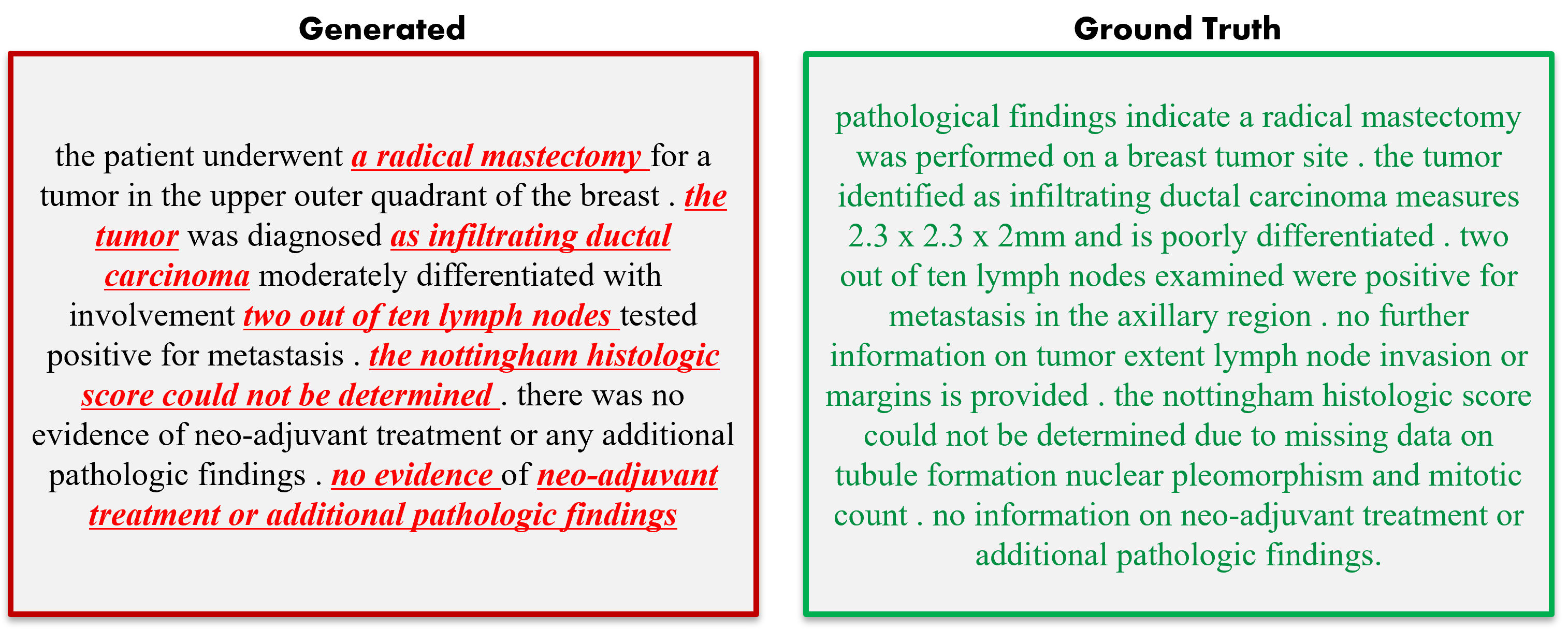} 
    \caption{\textbf{Qualitative comparison of generated and reference pathology report text.}
    Correctly generated words and phrases are highlighted in \textcolor{red}{red}.}
    \label{fig:gen_ref_text}
\end{figure*}

\subsection{Datasets}
We perform experiments on the PathText-BRCA dataset based on the data partitioning protocol described in~\cite{chen2024wsicaption}. The dataset is divided into training, validation, and test subsets with consistent split settings. To prevent information leakage, we remove duplicated patient samples across subsets, ensuring patient-level independence between splits. The final dataset includes 796 samples for training, 88 for validation, and 93 for testing.

\subsection{Implementation Details}
In our study, whole-slide images (WSIs) are processed using CLAM~\cite{lu2021data} to extract non-overlapping $256 \times 256$ tissue patches at 10$\times$ magnification. Visual representations are obtained using UNI~\cite{chen2024towards}, a foundation model pretrained on over 100 million tissue patches and more than 100,000 WSIs, which serves as the feature extractor.

Our architecture consists of 3 encoder layers and 3 decoder layers, each equipped with 4 attention heads and an embedding dimension of 512. Inspired by \cite{zhang2025historical}, we adopt the same hyperparameter configuration for the knowledge branch, where $k=0.4$, $m=20$, and $v=3$.

We train the model using the Adam optimizer with an initial learning rate of $1\times10^{-4}$ and a weight decay of $5\times10^{-5}$. During inference, beam search with a beam size of 3 is used for sequence generation. All experiments are conducted on a single NVIDIA A100 GPU (40GB).

\begin{table*}[t]
\centering
\caption{Ablation study on PathText (BRCA). BLEU-n denotes n-gram BLEU score. $\uparrow$ indicates higher is better.}
\label{tab:ablation_ranger}

\resizebox{\textwidth}{!}{%
\begin{tabular}{lcccccccccc}
\toprule
Model 
& Reranker 
& MoE 
& Noisy Top-$k$ 
& Load Balance 
& BLEU-1$\uparrow$ 
& BLEU-2$\uparrow$ 
& BLEU-3$\uparrow$ 
& BLEU-4$\uparrow$ 
& METEOR$\uparrow$ 
& ROUGE-L$\uparrow$ \\
\midrule

Baseline (cosine retrieval) 
& $\times$ & $\times$ & $\times$ & $\times$
& 0.4500 & 0.2960 & 0.1960 & 0.1350 & 0.1800 & 0.2930 \\

+ MLP Reranker 
& $\checkmark$ & $\times$ & $\times$ & $\times$
& 0.4543 & 0.2997 & 0.1993 & 0.1387 & 0.1837 & 0.2978 \\

+ MoE Decoder (sparse) 
& $\checkmark$ & $\checkmark$ & $\times$ & $\times$
& 0.4540 & 0.2989 & 0.1987 & 0.1380 & 0.1829 & 0.3005 \\

+ Noisy Top-$k$ Routing 
& $\checkmark$ & $\checkmark$ & $\checkmark$ & $\times$
& 0.4569 & 0.3019 & 0.2013 & 0.1410 & 0.1858 & 0.3022 \\

\midrule
\rowcolor{gray!60}
+ Load Balance ($\lambda=0.01$) \textbf{(Ours)} 
& $\checkmark$ & $\checkmark$ & $\checkmark$ & $\checkmark$
& \textbf{0.4598} 
& \textbf{0.3044} 
& \textbf{0.2036} 
& \textbf{0.1435} 
& \textbf{0.1883} 
& \textbf{0.3038} \\
\bottomrule
\end{tabular}%
}
\end{table*}

\subsection{Comparison between our methods and other report-generation baselines}

Table~\ref{tab:pathtext_brca_results} reports quantitative results on the PathText (BRCA) benchmark. 
Compared with earlier CNN-RNN and attention-based LSTM models, 
Transformer-based approaches consistently achieve stronger performance 
across BLEU and ROUGE metrics, highlighting the advantage of self-attention 
for modeling long-range dependencies in pathology reports.

Among prior approaches, BiGen~\cite{zhang2025historical} achieves the strongest performance, 
demonstrating the effectiveness of knowledge-enhanced report generation. 
Building upon this foundation, \textbf{RANGER} consistently surpasses BiGen across all metrics. 
Notably, RANGER improves BLEU-4 by +0.0085 and ROUGE-L by +0.0108, 
indicating improved higher-order linguistic coherence and more faithful alignment 
with ground-truth diagnostic narratives. Importantly, RANGER also improves METEOR and ROUGE-L, indicating enhanced semantic alignment and structural consistency 
with ground-truth diagnostic narratives. 
The significant improvements across BLEU, METEOR, and ROUGE-L vlidates that adaptive retrieval refinement and sparse expert 
specialization jointly enhance both lexical precision and semantic coherence in pathology report generation.

\subsection{Ablation Studies}
\label{subsec:ablation}

We conduct ablation experiments to evaluate the contribution of the proposed learned reranker (RR) and sparse MoE decoder. All ablations are performed on the PathText (BRCA) dataset.


Table~\ref{tab:ablation_ranger} presents a thorough abaltion study for each module we proposed. Based on experimental results, introducing the learned reranker alone leads to slight improvements on BLEU-4, METEOR, and ROUGE-L compared to the backbone, indicating that compatibility-based refinement enhances
fine-grained region-text alignment. Notably, the reranker contributes more consistent enhancements across metrics when combined with MoE decoder, demonstrating that refined semantic signals are more effective with high capacity decoder.

Replacing the standard FFN with the sparse MoE decoder consistently improves BLEU-1/2/3/4 and ROUGE-L, highlighting the benefit of token-level transformation specialization.
In particular, MoE enhances long-span linguistic consistency as reflected by higher BLEU-3 and BLEU-4 scores., which is hard to improve given the sample size of the dataset.

Combining both modules yields the best overall performance, demonstrating that the two-stage input-side semantic refinement
and output-side transformation specialization (MoE) address complementary aspects of pathology report generation.

\paragraph{Effect of Learned Reranking.}
\begin{table}[!t]
\centering
\caption{Stage-1 recall size $K$ (top-$k$=3 fixed).}
\label{tab:recall_k_single}
\resizebox{\columnwidth}{!}{%
\begin{tabular}{lcccccc}
\toprule
Recall Size $K$ & BLEU-1 & BLEU-2 & BLEU-3 & BLEU-4 & METEOR & ROUGE-L \\
\midrule
$K=10$ & 0.4568 & 0.3022 & 0.2018 & 0.1420 & 0.1868 & 0.3018 \\
\rowcolor{gray!45}
$K=20$ (Ours) & \textbf{0.4598} & \textbf{0.3044} & \textbf{0.2036} & \textbf{0.1435} & \textbf{0.1883} & \textbf{0.3038} \\
$K=50$ & 0.4576 & 0.3028 & 0.2023 & 0.1426 & 0.1873 & 0.3025 \\
\bottomrule
\end{tabular}%
}
\end{table}

\begin{table}[t]
\centering
\caption{Final top-$k$ selection ($K=20$ fixed).}
\label{tab:topk_single}
\resizebox{\columnwidth}{!}{%
\begin{tabular}{lcccccc}
\toprule
Top-$k$ & BLEU-1 & BLEU-2 & BLEU-3 & BLEU-4 & METEOR & ROUGE-L \\
\midrule
$k=1$ & 0.4561 & 0.3016 & 0.2012 & 0.1415 & 0.1862 & 0.3014 \\
\rowcolor{gray!45}
$k=3$ (Ours) & \textbf{0.4598} & \textbf{0.3044} & \textbf{0.2036} & \textbf{0.1435} & \textbf{0.1883} & \textbf{0.3038} \\
$k=5$ & 0.4580 & 0.3031 & 0.2025 & 0.1428 & 0.1875 & 0.3029 \\
\bottomrule
\end{tabular}%
}
\end{table}

To validate the effectivess of the two-stage retrieval module, we perform ablation study on the number of retrieved sentence embedding in stage 1 and number of refined sentence embedding in stage 2. The complete ablation study for top-k selection for stage 1 and stage 2 is shown in Table. \ref{tab:recall_k_single} and \ref{tab:topk_single}.
The reranker consistently improves higher-order BLEU scores
and ROGUE-L score, indicating that simple cosine retrieval from memory bank is insufficient compared to our two-stage refined semantic retrieval. For Stage-1 recall size K, we ablate on K=10, 20, and 50, and we observe that K = 20 have the best performance, demonstrating the need for efficient top-k selection for corase-grained selection in stage-1.

Table \ref{tab:topk_single} examines the effect of the final top-k selection in the two-stage reranker, with the stage-1 recall size fixed at K=20. Based on the results, with k = 1, the retrieved textual embedding have the lowest contextual diversity. When we increaase the number of k to 3, the softmax aggregation of 3 high-quality candidates provide richer contextual guidance. When k = 5, the performance slightly degrade, and this is potentially due to introduce more lower-ranked candidates introduces noise that might diluted the aggregated representations. These results illustrate that two-stage design, broad recall followed by fine-grained selection is much more efficient than simple aggregation-based technique, such as mean pooling over all possible candidates.

\paragraph{Effect of Sparse Expert Routing.} 

We further analyze the impact of MoE hyperparameters. Varying the number of experts shows that $E=4$ achieves the best BLEU-4 score, which is a thorough and challenging metric that measures 4-gram performance between generated report and ground truth report. Based on experiment results, larger expert counts (e.g., $E=8$)
lead to performance degradation, which indicates that excessive expert capacity hinder effective specialization during the report token generation process. Additionally, this behavior is likely due to limited training data, where excessive capacity results in under-utilized experts and unstable specialization.

We also investigate the effect of top-$k$ routing.
Preliminary results indicate that top-2 routing outperforms
top-1 routing, and top-3 routing, suggesting that collaborative expert activation better captures heterogeneous diagnostic reasoning patterns. This concur with the reasoning that sparsely gated ensemble learning enhances representation without introducing excessive redundancy. 

Finally, we evaluate different load-balancing coefficients $\lambda$.
Moderate regularization ($\lambda=0.01$) yields more stable training and improved performance compared to smaller values. Without any load-balancing loss, the noise-top-k router still enhances over the baseline model with regular FFN layer, which might lead to unequal token assignment and results in under-utilization of some experts. Based on the ablation results, we found that ($\lambda=0.001$) slightly decrease thet performance, indicating that insufficiently strong constraint destabalizes the routing distribution. Moderate regularization with ($\lambda=0.01$) significantly enhances the performance, as it sufficiently encouraging load balance while still allowing expert specialization, while increasing ($\lambda=0.1$) introduces excessive constraints that over-regularize routing decision will slightly suppress expert specialization, which is expected. Our results demonstrate that moderate load-balancing is crucial for stable MoE training on this dataset.

\begin{figure}[t]
\centering
\includegraphics[width=\columnwidth]{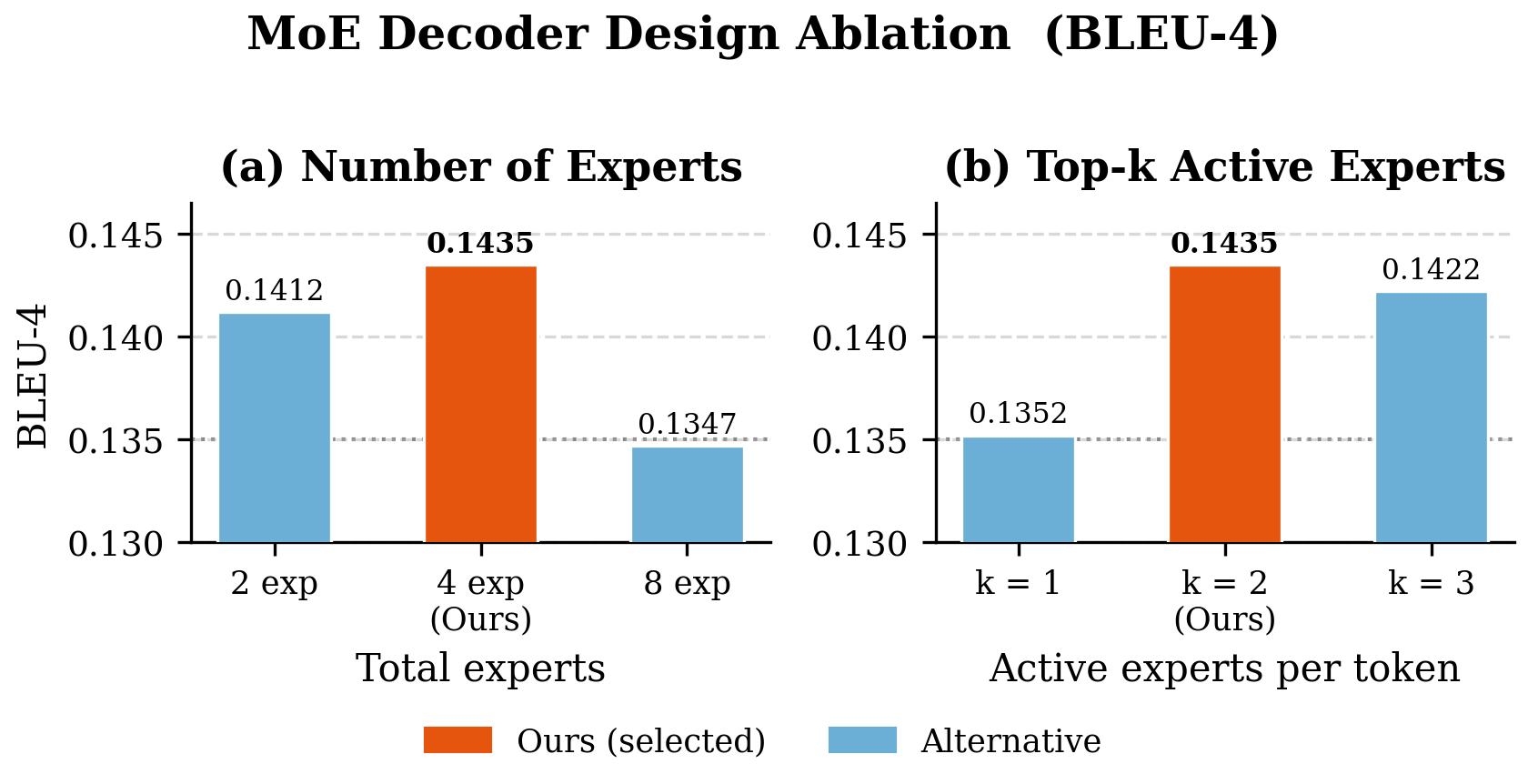}
\caption{
Ablation study on the Mixture-of-Experts (MoE) decoder. 
Left: Performance variation with different numbers of experts. 
Right: Effect of top-$k$ routing during expert activation. 
Moderate expert capacity with sparse routing achieves the best trade-off between specialization and stability.
}
\label{fig:lambda_ablation}
\end{figure}

\begin{figure}[t]
\centering
\includegraphics[width=\columnwidth]{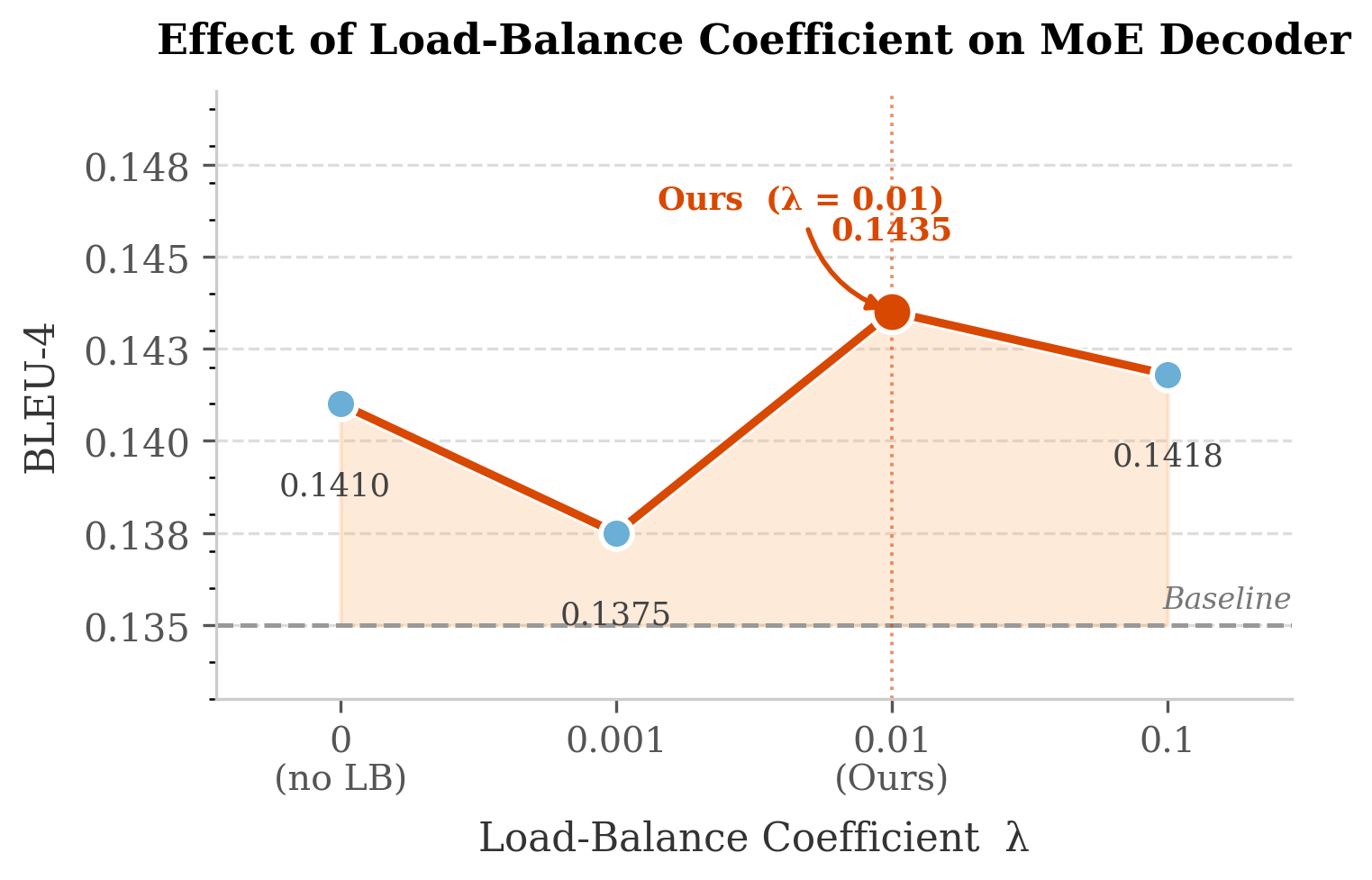}
\caption{
 Effect of load-balance coefficient $\lambda$ on MoE decoder performance (BLEU-4). $\lambda$=0.01 yields the best performance. Too-small $\lambda$ (0.001) fails to prevent expert collapse and performs worse than no load balancing, while too-large $\lambda$ (0.1) over-constrains routing. Dashed line: cosine-retrieval baseline
}
\label{fig:moe_ablation}
\end{figure}

\section{Conclusion}

In this work, we present RANGER, a novel framework for pathology report generation from whole-slide images. Our approach integrates visual feature extraction with a knowledge-enhanced generation mechanism to improve the quality and correctness of generated reports. We evaluate the proposed method on the PathText-BRCA dataset under a standardized experimental protocol and demonstrate competitive performance compared with existing report generation approaches.

The proposed framework effectively captures informative visual representations from high-resolution WSIs and leverages structured knowledge to guide report generation. These results suggest the potential of knowledge-enhanced modeling for improving clinically meaningful pathology report generation.

In future work, we plan to extend our framework to larger multi-institutional datasets and explore more advanced knowledge integration strategies to further improve generation quality and robustness.

{\small
\bibliographystyle{ieeenat_fullname}
\bibliography{main}
}


\end{document}